\documentclass[runningheads]{llncs}
\usepackage{listings}

\usepackage[backend=bibtex,style=numeric-comp,sorting=none]{biblatex} 

\usepackage{soul}

\usepackage{hyperref}
\usepackage[table]{xcolor}
\usepackage{graphicx}
\usepackage{geometry}
\usepackage[marginal]{footmisc}
\usepackage{enumitem}
\usepackage{authblk}
\usepackage{amsmath}
\usepackage{float}
\definecolor{LimeGreen}{rgb}{0.2, 0.8, 0.2}

\title{Solving the Extended Job Shop Scheduling Problem with AGVs -- Classical and Quantum 
    Approaches\thanks{The presented work was funded by the German Federal Ministry 
        for Economic Affairs and Energy within the project ``PlanQK'' 
        (BMWi, funding number 01MK20005)~\cite{PlanQK} and licensed under
        CC BY-NC-ND 4.0.}}
        
\titlerunning{Extended Job Shop Scheduling with AGVs -- Classical and Quantum Approaches}

\author{Dr. Marc Geitz\inst{1}\footnote{the authors are listed in alphabetical order.} 
\and Dr. Cristian Grozea\inst{2} 
\and Wolfgang Steigerwald\inst{3} \and Robin Stöhr\inst{4} \and \\ 
Dr. Armin Wolf\inst{5}\orcidID{0000-0003-3940-0792}}

\authorrunning{M. Geitz, C. Grozea, W. Steigerwald, R. Stöhr, and A. Wolf}

\institute{Telekom Innovation Laboratories \email{marc.geitz@telekom.de} \and
Fraunhofer FOKUS \email{cristian.grozea@fokus.fraunhofer.de} \and
Telekom Innovation Laboratories \email{wolfgang.steigerwald@telekom.de} \and
Telekom Innovation Laboratories \email{rb.stoehr@gmail.com} \and
Fraunhofer FOKUS \email{armin.wolf@fokus.fraunhofer.de}}

\begin{document}

\maketitle

\begin{abstract}
The subject of Job Scheduling Optimisation (JSO) deals with the scheduling of jobs in an organization, so that the single working steps are optimally organized regarding the postulated targets. In this paper a use case is provided which deals with a sub-aspect of JSO, the Job Shop Scheduling Problem (JSSP or JSP). 

As many optimization problems JSSP is NP-complete, which means the complexity increases with every node in the system exponentially. The goal of the use case is to show how to create an optimized duty rooster for certain workpieces in a flexible organized machinery, combined with an Autonomous Ground Vehicle (AGV), using Constraint Programming (CP) and Quantum Computing (QC) alternatively. The results of a classical solution based on CP and on a Quantum Annealing model are presented and discussed. All presented results have been elaborated in the research project PlanQK.

\end{abstract}


\section{Introduction}
\label{sec:overview}

\subsection{Problem Motivation and Description}
The calculation of duty rosters is, since many years, a special subject in the departments of Operation Research. JSSP is a problem which is NP-hard, which means that with each additional node in the manufacturing sequence the solution space expands exponentially. But not only the number of nodes increases the complexity, there are further constraints to be considered like:
\begin{itemize}[nosep]
	\item Functionalities of the machines, i.e. not every machine can realize each production 
	step.
	\item Processing times for specific work items.
	\item Machinery set-up times. 
	\item Changeover times for switching tools on a machine.
	\item Temporary storage times of the workpieces. 
\end{itemize}

The prioritization of workpieces for special customers or the prioritization of workpiece groups will increase the problem complexity even further.

Individualized transports of workpieces on the shop floor within the paradigms of Industry 4.0 increase the complexity even more. For this reason we deduced a use case including most of the mentioned  constraints but small in scale, however, with the possibility to scale up later.

\section{Use Case Scenario}

For the initial scenario we consider four machines with different tool settings. Autonomous Ground Vehicles (AGVs) are used to transport the workpieces on the shopfloor.

\subsection{Constraints and Definitions}

\begin{figure}[t]
  \includegraphics[width=\linewidth]{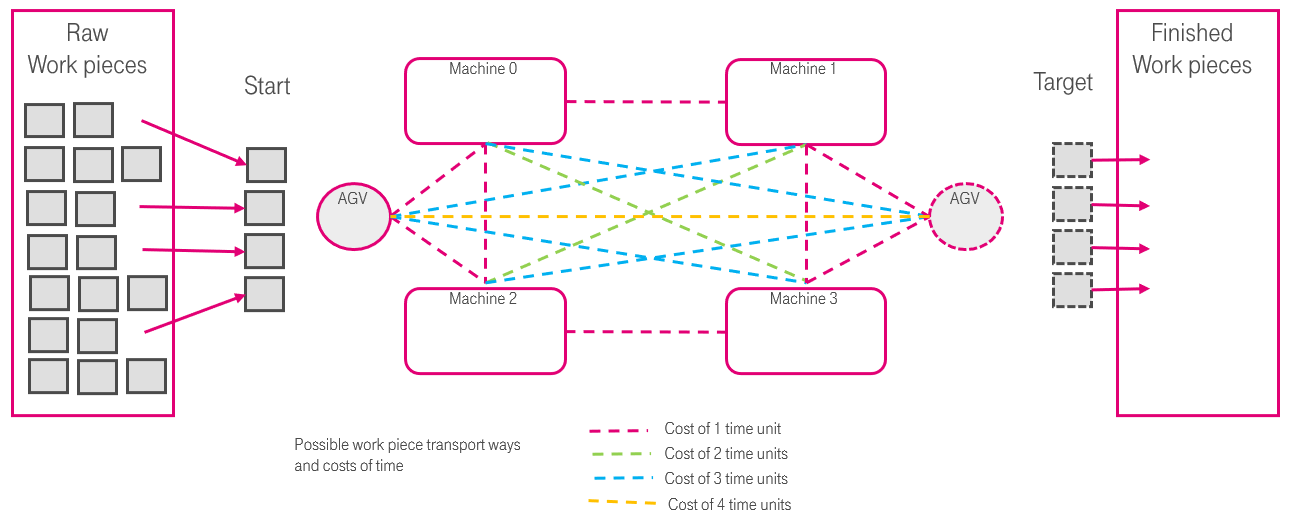} 
  \vspace{-1.0em}
  \caption{Initial use case scenario}
  \label{fig:JSO1-1}
\end{figure}

\begin{itemize}[nosep]
	\item Each machine has a specific functionality which cannot be performed by another machine.
	\item Each machine has an unlimited storage before and after each processing step.
	\item A processing step (also named here ``working step'' or ``operation''), as well as 
	on the machine as on the transport, is non-preemptive. 
	\item At most one workpiece (or ``operation'') can be processed on a machine at any time.	
	\item Each working step should have the same priority.
	\item For each working step the processing sequence and the duration is known in advance.
	\item For each workpiece the working steps must be processed in a fixed sequence, named ``job''.
    \item Every operation must run exactly once.	
    \item Every operation inside a job can only start if the previous one is finished. 	 	
	\item The transportation of the workpieces is done by an AGV.
	\item The AGV and all the workpieces are positioned at the beginning on the start point (``Start'') and afterward’s on the target point (``Target''). 
	In some scenarios there can be several start and target points.
	\item The AGV can carry at most one workpiece on each transportation step, between ``Start'', machinery and ``Target''.
	\item The required transit time between ``Start'', machinery and ``Target'' must be incorporated during the planning and optimization of the duty roster.
	\item The durations of the transit times between the ``Start'', machinery and ``Target'' are known in advance.
	\item The total processing time should be minimized. The total processing time is the time consumed for all parts being manufactured, according to the duty roster, starting at``Start''and ending up at ``Target''.
\end{itemize}
An initial scenario is presented schematically in Figure~\ref{fig:JSO1-1}.

\subsection{Job-Shop Scheduling Problem}

\subsubsection{Problem definition}
We have a set of $N$ jobs $J =\{j_0,...,j_{N-1}\}$. Each job consists of a sequence of operations,
where each operation is to be executed on one of the $M$ machines $\{m_0,...,m_{M-1}\}$ and will take a known time~$t$.

With a given input of jobs, operations, and machines, as well as the number and initial positions of the AGVs, the task is to find an optimal schedule, that will minimize the time~$T$ needed to complete all operations \cite{google_job_shop}.

\section{Related Work}

In optimization problems, we are trying to find the best solution from all feasible solutions. We can convert those problems into energy minimization problems and use the effects of quantum physics to find the solution of the lowest potential energy~\cite{dwave_introduction}.
In quantum annealing we have, instead of classical bits, so-called \emph{qubits}, which on top of being in a state of 0 or 1, those can also be in a superposition, which corresponds to both states at the same time. 

Quantum Annealing (QA) assumes a system in full superposition being in its lowest energy state. The system is then slowly (adiabatically) conveyed to an energy function describing the optimization problem - the adiabatic theorem \cite{adiabatic_theorem} assures that the system will remain on its lowest energy state during the transition. At the end of the annealing process, the system has turned from a fully superimposed quantum state to a classical state, where the qubits can easily be read out.

The Job Shop Scheduling problem (JSP) has been solved using the technique of simulated annealing \cite{YamadaTNakanoR1996,chakrabortyJobShopScheduling2013,ZhangRui2013}. A quantum annealing solution has first been introduced in~\cite{venturelli2015quantum} where the JSP has been modelled as cost or energy function to be minimized (QUBO, see below). A basic implementation for a DWave quantum annealer is presented in~\cite{dwaveexamples}. The modelling or implementation of a more complicated JSP problem, especially with the introduction of AGVs or robots to transport goods between machines on the shop floor is to the best of 
our's knowledge not yet known. 

A \emph{quadratic unconstrained  binary optimization problem} (QUBO) is a minimization problem of this form:
\begin{equation}
	\min_{x\in\{0,1\}^N}{\sum_{i} Q_{i,i}x_{i}+\sum_{i<j}Q_{i,j}x_{i}x_{j}}\label{eq_qubo}
\end{equation}
where $x=(x_i)_{i=1\dots N}$ are binary variables and the real-valued matrix $Q$ of size $N\times N$ fully defines the QUBO instance.
The matrix Q is by convention upper diagonal. Thanks to the similarity of the QUBO formula to the Ising Model \cite{dwave_qubo}, QUBO matrices can be transformed trivially into Ising Hamiltonian matrices and thus solutions for the QUBO problems can be produced by quantum annealers like those of DWave.
The constraints of the problems to be modelled as QUBOs are integrated into the objective function as additional terms that achieve their minimum only when the respective constraints are satisfied.
A quantum computer will compute the minimum of the QUBO function, hence solve the optimization problem with all constraints fulfilled.


\emph{Constraint Programming} (CP) deals with discrete decision an optimization problems. This paradigm combines pruning algorithms known from Operations Research (OR) reducing the search space and heuristic tree search algorithms, e.g. instances of depth-first search, known from Artificial Intelligence (AI) in order to
explore the search space. The advantage of CP is that each decision during search will be used to 
further restrict the search space, i.e. pruning branches in the search tree. Using specialized OR
pruning algorithms, CP is the appropriate approach for solving scheduling
problems~\cite{pBaptiste:etal:01} even those considered here. Compared to local search approaches, e.g., \emph{Simulated Annealing} or \emph{Quantum Annealing} CP offers several advantages: 
\begin{itemize}
    \item The solutions found (schedules) always satisfy the specified constraints,
    which is not the case in local search, where the violation of the constraints is only minimized. 
    \item CP can guarantee the quality of intermediate solutions during the optimization process.
    \item The optimality of the best solutions can be achieved and proven with complete search.
\end{itemize}
In CP there exist highly efficient methods for reducing the finite value ranges of the decision 
variables (start times, durations, end times, possible resources etc.) of activities on 
exclusively available (\emph{unary}) resources. Some of those are: \emph{overload checking}, 
\emph{forbidden regions}, \emph{not-first/not-last detection}, \emph{detectable precedences}
and \emph{edge finding}~\cite{baptisteEdgeFindingConstraintPropagation1996,vilimLogFilteringAlgorithms2004,wolfPruningSweepingTask2003}.
Some recent version of those pruning methods support \emph{(sequence-dependent) transition times} 
and pruning for 
\emph{optional activities}~\cite{wolfConstraintBasedTaskScheduling2009,vancauwelaertEfficientFilteringAlgorithm2020}. 
The optional activities (that can be optionally scheduled, but not necessarily) are the basis for certain modelling approaches for scheduling activities on \emph{alternatively exclusive resources}~\cite{wolfRealisingAlternativeResources2005}.
Such activities are in our case the transport tasks, that can be scheduled on alternative available AGVs.
However these activities change the locations of the AGV which has to be respected: 
Transition times occur between two consecutive transport tasks 
when the destination of the first transport task differs from the origin of the second one. 
In those cases the used AGV has to travel (empty) from one location to another. 

Optimization in CP is generally done by \emph{Branch-and-Bound} (B\&B), i.e. through an 
iterative search for ever better solutions. The optimization follows either a \emph{monotonous} 
or a \emph{dichotomous} bounding strategy combined with a mostly problem specific one,
generally full depth search~\cite{hofstedtEinfuehrungConstraintProgrammierungGrundlagen2007}.
This combined strategy (B\&B with full depth search) guarantees that the best solutions are found
and their optimality is proven. Furthermore, dichotomous bounding allows to quantify the quality of
already found, intermediate solutions.
Recent advances in CP follow a hybrid approach that combines search space pruning with the satisfiability problem solving (SAT Solving) for classical propositional logic~\cite{stuckeyLazyClauseGeneration2010}. 
There, clauses are generated during constraint processing (propagation) and further processed by SAT Solving~\cite{feydyLazyClauseGeneration2009}. Promising results show that this hybrid 
approach is ideally suitable for solving scheduling problems in a classical way~\cite{schuttExplainingTimeTableEdgeFindingPropagation2013,schuttExplainingCumulativePropagator2011,schuttSolvingRCPSPMax2013,schuttExplainingProducerConsumer2016}.

\section{Methods}

\subsection{Solving the Extended JSP with AGV Transport using Constraint Programming}\label{subsec:CP}

An AGV performs transport tasks in ${\cal T} = \{t_1, \ldots, t_n\}$ between some locations
in~${\cal L} = \{L_1, \ldots, L_m\}$. If the destination and the departure locations between two
consecutive transport tasks are different some additional transitions have to be performed. 
Let $d(P, Q)$ be the \emph{travel time} between any two locations~$P, Q \in {\cal L}$. 
Then it is assumed that
\begin{eqnarray*}
	d(L, L) &  =  & 0 \qquad\mbox{for any location~$L \in {\cal L}$,}\\
	d(P, Q) & \ge & 0 \qquad\mbox{for any two different locations~$P, Q \in {\cal L}, P \ne Q$ and}\\
	d(P, Q) & \le & d(P, R) + d(R, Q) \qquad\mbox{for any locations~$P, Q, R \in {\cal L}$.} 		
\end{eqnarray*}
The last inequality is also known as \emph{the triangle inequality} and satisfied in Euclidean spaces.

Each transport task~$t_i \in {\cal T}$ is annotated by its \emph{departure location}~$P_i \in {\cal L}$
where the AGV picks up a workpiece and the \emph{destination location}~$Q_i \in {\cal L}$ where the 
AGV releases a workpiece, i.e. let~$t_i = t_i(P_i,Q_i)$. Further, let~$s_i = s_i(P_i)$ be 
\emph{the (variable) start time at departure location~$P_i \in {\cal L}$} of the transport
task~$t_i(P_i,Q_i)$ and $e_i = e_i(Q_i)$ be \emph{the (variable) end time at destination 
location~$Q_i \in{\cal L}$} 
of this task. Further let~$d_i = d(P_i,Q_i)$ be the \emph{duration} of task~$t_i(P_i,Q_i)$
determined by the travel time between its locations so that 
\begin{eqnarray}
	s_i(P_i) + d(P_i, Q_i) = e_i(Q_i) && \mbox{holds for $i=1, \ldots, n$.}
\end{eqnarray}

The transport tasks in~${\cal T}$ of an AGV must be scheduled in \emph{linear order}; i.e. 
${\cal T}=\{t_1, \ldots, t_n\}$ must be ordered in such a way that 
\begin{eqnarray} 
  e_i(Q_i) + d(Q_i, P_{i+1}) \le s_{i+1}(P_{i+1}) && \mbox{holds for $i=1, \ldots, n-1$.} 
  \label{eq:setup}
\end{eqnarray}   
There, $d(Q_i, P_{i+1})$ is the \emph{sequence-dependent transition time} between the destination 
location~$Q_i$ of the $i$-th transport task and the departure location~$P_{i+1}$ of the successive
$(i+1)$-th transport task.

In order to satisfy these sequence-dependent transition times on exclusively available resources we
adapted the approach presented in~\cite{wolfConstraintBasedTaskScheduling2009}. Therefore
additional tasks to be scheduled on additional resources -- one for each location -- are 
considered: For each ``real'' transport task~$t_i(P_i,Q_i)$ and each location~$L$ we consider an
additional ``virtual'' transport tasks ${t'}_i^L$ with start and end times
\begin{eqnarray*}
    {s'}_i^L & = & s_i(P_i) + d(P_i, L) \\
	{e'}_i^L & = & e_i(Q_i) + d(Q_i, L)
\end{eqnarray*}
The duration of each additional task ${t'}_i^L$ namely ${e'}_i^L - {s'}_i^L =  e_i(Q_i) + d(Q_i, L)
- s_i(P_i) - d(P_i, L) = s_i(P_i) + d(P_i, Q_i) + d(Q_i, L) - s_i(P_i) - d(P_i, L) = d(P_i, Q_i) +
d(Q_i, L) - d(P_i, L)$ is always non-negative due to the triangle inequality, i.e. each 
additional task is well-defined.

\begin{figure}[h]
  \includegraphics[width=\linewidth]{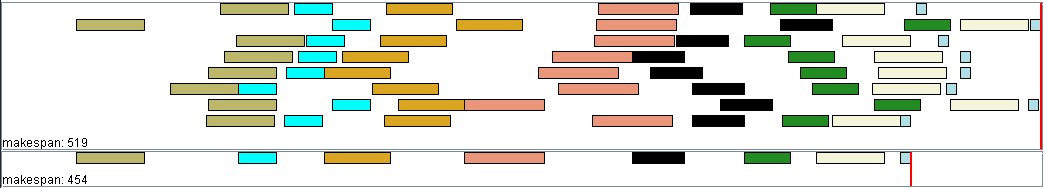}
  \vspace{-1.0em}
  \caption{The interrelationship between ``real'' transport tasks (bottom line) and their 
  according ``virtual'' tasks (upper part -- one line for each location) scheduled in the same
  linear order.}
  \label{fig:transitiontime}
\end{figure}

For an AGV with more than two transport tasks we proved that $e_i(Q_i) + d(Q_i, P_j) \le s_j(P_j)$
holds for any two different transport tasks $t_i,t_j \in {\cal T}, t_i \ne t_j$ 
if and only if ${e'}_i^L \le {s'}_j^L$ holds for each location~$L \in {\cal L}$.\footnote{Details on 
the proof of this statement can be found in~\cite{wolfConstraintBasedTaskScheduling2009}.}
This means that scheduling the additional tasks ${t'}_i^L$ in the same linear order for each
location~$L \in {\cal L}$ results in a schedule where the transport tasks are scheduled in the same 
order respecting the necessary transition times between different locations as shown
in Figure~\ref{fig:transitiontime}. There each task~$t_i$ and the according tasks~${t'}_i^L$
have the same color.

For scheduling the transport tasks of the AGVs as well as the tasks on the machines
we used the Fraunhofer FOKUS developed Constraint Programming library {\tt 
firstCS}~\cite{wolfFirstCSNewAspects2012} 
to model the Job Shop Scheduling problems including AGVs for transport. The {\tt firstCS} library
includes implementations of algorithms for \emph{unary} resources adapted 
from~\cite{vilimLogFilteringAlgorithms2004,vilimUnaryResourceConstraint2004,%
wolfPruningSweepingTask2003}. We handled the ordering of tasks globally using 
an adapted version of \emph{global difference constraints} ~\cite{feydyGlobalDifferenceConstraint2008}. 
The sequence-dependent transition times were handled on algorithmic level using a new approach 
described in~\cite{vancauwelaertEfficientFilteringAlgorithm2020}. These filtering algorithms are
pruning the search space and are used together with a depth-first tree-search called 
\emph{domain reduction}~\cite{pGoltz95} for minimizing the makespan. 
The chosen algorithms also support \emph{optional activities} so that they are 
applicable in the case where transport tasks can be performed by more than AGV. 
In this general case, the algorithms presented in~\cite{wolfRealisingAlternativeResources2005} 
are used to deal with such alternative exclusive resources.

\subsection{Solving the Extended JSP with AGV Transport on a Quantum Annealer}

We model our problem as a \emph{quadratic unconstrained binary optimization} problem (\emph{QUBO}, cf. (\ref{eq_qubo})). 

\subsubsection{Variable definitions and notations}

Let  $T$ be the planning horizon and $I$ be the number of machine operations. 
Starting with the variable structure used in \cite{venturelli2015quantum}, we index our variable array $x$ by operation index $i$ and running time $0\le t< T$:
$x_{i,t} = 1$ if and only if operation $O_i$ starts at time $t$,
where $O_i$ denotes operation number $i$. 
The operations are ordered in a lexicographical way, starting from the first operation of the first job, going to the last operation of the last job. 

Additionally we denote the processing time of operation $i$ with $p_i$, and with $k_n$ the index of the last operation of the job $j_n$.
Let $W=(w_{a,b})$ be the distance matrix, where $w_{a,b}$ stores the distance from machine $a$ to machine $b$.

\paragraph{Walking Operations}
To include the AGV movements into the model, we introduce walking operations in addition to the standard (machine) operations. Between every two successive standard operations inside of a job, we add $B$ walking operations, where $B$ denotes the number of AGVs. 
The individual walking operations are denoted with $\xi$. All $B$ walking operations between two standard operations are combined in a set that we name ``the parent walking operation''. One of these operations, denoted with $\Psi$, will have the structure $\Psi_x=(\xi_{x0}, \xi_{x1}, ...)$. Out of every parent walking operations, one individual one must be picked, to transport the piece. Individual walking operations are defined as a tuple $\xi_x=(\text{b}, w_{m_{s}m_{e}}, m_s, m_e)$ where $b$ is the AGV executing this operation, $m_s$ is the start machine the workpiece needs to be picked up from and $m_e$ is the end machine the piece needs to be delivered to.
Those operations are appended at the end of the standard operations. 
\paragraph{Predecessor Walking Operation}
Additionally we define $\Omega_i$ as the predecessor walking operation, of the standard operation $i$. This contains the walking operation of all AGVs, that come directly before $i$. 
For those indices $i$ for which $O_i$ is the first operation of a job, $\Omega_i$ is an empty set.

\paragraph{Predecessor Standard Operation}
We define $\Phi_i$ as the standard operation, previous to the walking operation $i$. 
$I_m$ is defined to be the set of all machine operations executed on machine $m$. 
$I_b$ is defined to be the set of all walking operations executed by the AGV $b$. 
$I_\Psi$ is the set of all operations in the parent walking group $\Psi$.

\subsubsection{Constraints}
We need to account for the following constraints: 

\paragraph{h1}
\underline{\emph{Every machine operation must run exactly once}}. With a planning horizon $T$, our variables for operation $O_i$ would consist of $x_{i,0}, x_{i,1},\dots, x_{i,T-1}$. Here we need to enforce, that exactly one of the three variables is $1$, and the rest $0$. That means we want to increase the penalty, when less than, one or more than one of them is $1$. We can do that by iterating over the set of all operations $I$, and for each $i \in I$ we subtract $1$ from the sum off all $t \in T$, and square the result. This way the penalty term for this constraint is $0$ if exactly one operation is picked, and higher than that otherwise, which will enforce exactly one operation being picked.
\begin{equation}
	h_1(\bar{x}) = \sum_{i=0}^{I-1} \left(\sum_{t=0}^{T-1} x_{i, t}-1\right)^2
	\label{eq_h1}
\end{equation}

\paragraph{h2}
\underline{\emph{Every machine can only work on one operation at a time}}. 
Here, we define a set $R_m$ that contains all indices of operation $i$ with its corresponding starting time $t$ and a second operation $i'$ with its time $t'$ for which scheduling them as such leads to an overlap on the same machine:
\begin{equation*}
	R_m = \{(i,t,i',t'):(i,i') \in I_m \times I_m, i \neq i', 0 \leq t, t' \leq T, 0 \leq t'-t < p_i \}
\end{equation*}

\begin{equation}
 h_2(\bar{x}) = \sum_{m=0}^{M-1} \sum_{(i,t,i',t')\in R_m} x_{i,t}x_{i',t'} \label{eq_h2}
\end{equation}

\paragraph{h3}
\underline{\emph{Every operation inside a job can only start if the previous one is finished}}. For this constraint, we need to iterate over all jobs $n$, and pick the operations $i$ and execution times $t$ and $t'$, for which the sum of the processing time of $i$ with its execution time $t$ is larger than the execution time of the next operation $t'$, and therefore violates the constraint. 
\begin{equation}
	h_3(\bar{x}) = \sum_{n=0}^{N-1}\sum\limits_{\substack{t,t' \\ i,i+1\in I_m \\  t+p_i>t'}} x_{i,t}x_{i+1,t'} \label{eq_h3}
\end{equation}

\subsubsection{Objective Function}
The objective function is built by introducing a set of "upper bound" variables $U_t$ such that exactly one of them is true. Together they model a virtual finite domain variable $V$ with the domain $0\dots T$, such that $V=t$ if and only if $U_t=1$, i.e. $\sum_{t=0}^{T}{U_t}=1$.

\paragraph{h4}
\ul{\emph{Exactly one of the variables $U_t$ is set.}}
Enforcing 
this is done by adding the fllowing quadratic constraint to the target function:
\begin{equation} 
h_4(\bar{U})=(\sum_{t=0}^{T}{U_t}-1)^2 \label{eq_o2} 
\end{equation} 

\paragraph{h5} 
\ul{\emph{No planned operation ends after the upper bound.}}
The index $t$ where $U_t=1$ is understood to be an upper bound of the whole schedule, therefore we impose that
no $i,t_1,t_2$ exist such that
$x_{i,t_1}=1$, $U_{t_2}=1$ and $t_1+p_i>t_2$ (no planed operation ends after the upper bound).
Enforcing this condition can be done by adding this penalty term to the target function
\begin{equation}
h_5(\bar{x},\bar{U})=\sum_{\substack{i,t_1,t_2 \\ t_1+p_i>t_2}}{x_{i,t_1}U_{t_2}}\label{eq_o1}
\end{equation}

With those, we can introduce the core of the objective function
\begin{equation}
o(\bar{U})=\sum_{t=0}^{N}{t U_t} \label{eq_o3}    
\end{equation}
Its meaning is fairly straightforward: by minimizing $o(\bar{U})$ one gets a lowest upper bound that is also minimum amongst all consistent solutions (solutions satisfying all constraints) thus the solution that guarantees the shortest makespan.

\paragraph{Current target function} The constraints \eqref{eq_h1}\eqref{eq_h2}\eqref{eq_h3} combined with the objective functions constraints and components \eqref{eq_o2}\eqref{eq_o1}\eqref{eq_o3} give the basic JSP QUBO target function, the minimization of which produces lowest makespan solutions that satisfy all constraints defined so far.
\begin{equation}
	H_{T}(\bar{x},\bar{U})=\alpha h_{1}(\bar{x})+\alpha h_{2}(\bar{x})+\alpha h_{3}(\bar{x})+\alpha h_4(\bar{x},\bar{U})+\alpha h_5(\bar{U})+o(\bar{U})
\end{equation}

\subsubsection{Further Constraints}
We need to add four new constraints, to include the walking time between machines.
Once we add these constraints to our already existing target function, the QUBO solutions represent optimal schedules with regards to the walking time between the machines, and the number of AGVs we have.

\paragraph{h6}
\ul{\emph{For each transport between two standard operations, exactly one walking operation must be picked}}. 
Here we also make sure that for all operations belonging to the same parent walking operation and for all times, there is only one variable $x_{i,t}$, that will have the value 1, which means one and exactly one of the AGVs is selected to perform that transport.
\begin{equation}
	h_6(\bar{x}) = \sum_{\substack{j=0\\ \Omega_j\ne \emptyset}}^{I-1}\left(\sum_{i \in \Omega_j}\sum_{t=0}^{T-1} x_{i,t}-1\right)^2
\end{equation}

\paragraph{h7}
\underline{\emph{AGVs must have enough time to switch to the next machine}}. Here we iterate over all walking operations of each AGV, denoted as $I_b$. For all possible walking operations of these AGVs, we make sure that the starting time of operation $i' \in I_b$, is not before operation $i \in I_b$ is finished, plus the time the AGV needs to switch machines. In here, $me(i)$ defines the machine end of operation $i$ or the fourth element in the tuple. $ms(i')$ defines the machine start of operation $i'$, or the third element in the tuple.
\begin{equation}
	h_7(\bar{x}) = \sum_{b=0}^{B-1} \sum\limits_{\substack{i, i' \in I_b\\ i\ne i'}} \sum\limits_{\substack{t, t'\\ t`>t}} \left( x_{i,t}x_{i', t'}max\left(t+p_i+w_{me(i)ms(i')}-t',0\right)\right)
\end{equation}

\paragraph{h8}
\label{h8}
\ul{\emph{A standard operation can only be scheduled at time $t$, if the corresponding walking operation is finished before or at $t$}}. Using our previously defined \emph{Predecessor Walking Operation}, we iterate through all standard operations and their corresponding walking operations, and apply a penalty if the walking operation $i'$ is finished after the time $t$ of the standard operation.
\begin{equation}
	h_8(\bar{x}) = \sum_{i=0}^{I-1} \sum_{i' \in \Omega_i}\sum_{t, t'} x_{i,t}x_{i', t'}max\left(t'+p_{i'}-t,0\right)
\end{equation}

\paragraph{h9}
\ul{\emph{A walking operation can only be picked at time $t$, if the corresponding standard operation is finished before or at $t$}}. This will be the inverse of \nameref{h8}. Using the \emph{Predecessor Standard Operation}, we iterate through all walking operations and their corresponding standard operations and apply a penalty if the the previous standard operation starting at time $t'$ with it's individual processing time, is not finished before or at $t$.

\begin{equation}
	h_9(\bar{x}) = \sum_{j=0}^{I-1}\sum_{i\in \Omega_j}\sum_{t, t'} x_{i,t}x_{\Phi_i, t'}max \left(t'+p_{\Phi_i}-t,0\right)
\end{equation}

\subsubsection{Final QUBO.}

We add the new constraints to the previous target function, to get the final one for the extended JSP problem.
\begin{equation}
	H_{T}(\bar{x},\bar{U})=\alpha\sum_{i=1..9}h_{i}
	+o(\bar{U})
\end{equation}

\section{Results}

\subsection{Experimental Results Using the Constraint Programming Approach}
\begin{figure}[ht]
    \centering
    \includegraphics[width=\linewidth]{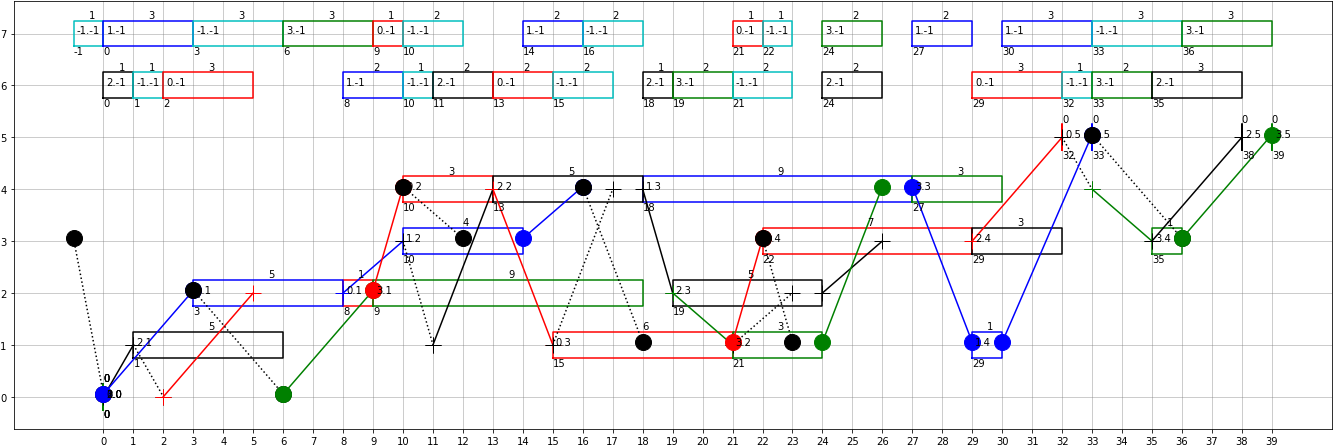}
    \vspace{-1.0em}
    \caption{Set6 -- optimal solution with makespan 39.
    Each of the colors red, green, blue, black corresponds to one of the workpieces. 
    The first two rows of horizontal blocks denote the transport operations, colored with the corresponding color of the corresponding workpiece. Empty transports are denoted with a supplementary color.
    The first row shows the transports performed by the first AGV, the second row the ones by the second AGV.
    The four rows that follow show the processing tasks scheduled on the four processing machines,
    again the color indicates the workpiece being processed.
    The tilted lines show the transport of the workpieces between the two special points ``Start''
    and ``Target'' and from one machine to another.}\label{fig:set6sol}
\end{figure}

\begin{table}[h]
\caption{Empirical results based on the {\tt firstCS} implementation}
    \label{tab:firstCS}
  \centering
    \begin{tabular}{r|c|c|c|r|r|r}
      Id & size & GD & B\&B & makespan & run-time & proof of opt. \\ \hline
      Set0 & 2x3x5x1 & $+$ & D & {\bf 9} & 0.3 sec. & { 0.3} sec.\\ \hline 
      Set1 & 4x4x24x1 & $+$ & D & {\bf 58} & 41 sec. & { 64} sec. \\ \hline 
      Set2 & 4x8x24x1 & $+$ & D & {\bf 45} & 87 sec. & { 87} sec. \\ \hline 
      Set3 & 4x4x24x1 & $+$ & D & {\bf 58} & 60 sec. & { 78} sec. \\ \hline 
      Set4 & 4x4x21x1 & $+$ & D & {\bf 49} & 18 sec. & { 30} sec. \\ \hline 
      Set5 & 4x4x21x1 & $+$ & D & {\bf 49} & 17 sec. & { 30} sec. \\ \hline 
      Set6 & 4x4x24x2 & $+$ & D & {\bf 39} & 106 sec. & { 1218} sec. \\ \hline 
      Set7 & 10x10x120x1 & $-$ & M & 412 & 692 sec. & --- \\ \hline 
      Set8 & 10x10x120x2 & $-$ & M & 224 & 200 sec. & --- \\
    \end{tabular}
\end{table}

The run-time experiments shown in Table~\ref{tab:firstCS} with the CP-based {\tt firstCS} 
implementation (cf.~Section~\ref{subsec:CP}) were performed on a Pentium i7-PC running Windows~10
with Java 1.8. The first column contains the problem identifiers, the second column the problem sizes
consisting of \#machines x \#jobs x \#tasks x \#AGVs. The third column shows whether the
\emph{global difference constraints} (GD) are used ($+$) or not ($-$). 
The forth column shows whether dichotomous (D) or monotonous (M) branch-and-bound (B\&B) was 
performed for minimizing the makespans shown in the fifth column (optimal makespans are in bold).
The next-to-last column shows the run-times to find the schedules with the presented makespans. 
The last column shows the total run-times for proving the optimality of found best solutions, 
i.e., including exhaustive search for even better solutions. 

Branch-and-bound optimization is based on different kinds of problem-specific, depth-first search: For the rather small, single AGV problem instances (Set0--Set5) the transport tasks on the AGV are
sequentially ordered first, then the the start times of all tasks are determined by successively 
splitting the domains of the start time variables in "earlier" and "later" subsets starting with the
smallest domain. For problem instances with two AGVs (Set6 and Set8) the transport tasks are
assigned to the AGVs beforehand. For the larger problem instances (Set7 and Set8) the ordering of
transport tasks on the AGVs is omitted. Further, branch-and-bound optimization starts with two trivial
bounds of the makespan: the sum of all transport durations between the machines divided by the number 
of AGVs as an initial lower bound and the sum of all task durations plus the longest travel time 
between all machines as an initial upper bound.

\subsection{Experimental Results Using the Quantum Approach}

On the problem "Set6", with two AGVs, 4 workpieces, 4 machines, and a total of 6 positions including``Start'' and``Target'' points which led to a QUBO matrix with 5525 qubits we have obtained the results shown in Table~\ref{tab:results}. The optimal solution, obtained using {\tt firstCS} is shown in Figure~\ref{fig:set6sol}. Optimising the QUBO with tabu search on CPU produced a similar solution of equal quality.

\begin{table}
    \caption{Empirical comparison of the results on the same problem ("Set6") on various quantum, quantum inspired and classical platforms. The quantum approaches processed the same 5525x5525 QUBO matrix.}
    \label{tab:results}
    \centering
    \begin{tabular}{l|r|c}
         Method & Time & Solution Quality (Makespan) \\
         \hline
         QC DWave Leap Hybrid v2 (5000 qubits) & 60 min & 57 \\
         Quantum-inspired Fujitsu DAU (8192 sim qubits) & 15 min & 53 \\
         CPU-based CBO {\tt firstCS} (single core) & {\bf\boldmath $< 2$ min} & \textbf{39} \\
         CPU Tabu Search (32 cores) & 120 min & \textbf{39}\\
    \end{tabular}
\end{table}

\section{Discussion and Conclusion}

\begin{figure}[H]
    \centering
    \includegraphics[width=\linewidth]{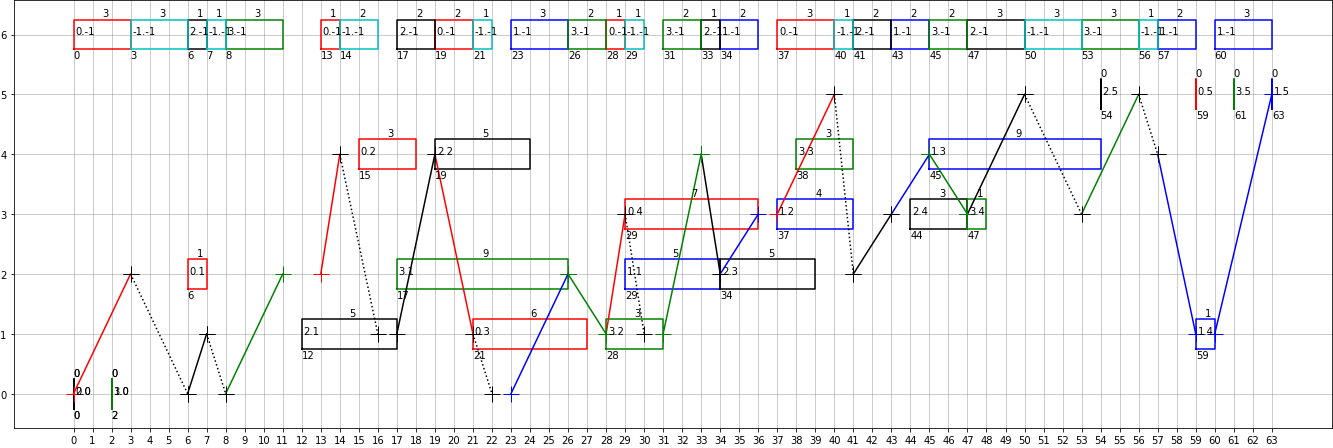}
    \vspace{-1.0em}
    \caption{Set1 - A tabu search solution with makespan 63 containing gaps, e.g. at time unit
    11. It can be compacted to a solution wit makespan 58 (optimal) by simply eliminating the gaps.}
    \label{fig:set1sol}
\end{figure}

An interesting behavior of all optimizers applied on the QUBO matrices is the presence of obviously unnecessary {\em gaps} in the produced solutions.
We have found out that by simply eliminating those gaps by doing any action as soon as the prerequisites are fulfilled, one obtains clearly better solutions, sometimes even the optimal ones. An example is given in Figure~\ref{fig:set1sol}.

In order to exclude the unlikely possibility that we made some error when building the QUBO matrices, we cross-validated the QUBO matrices $Q$ we have built against the solutions obtained with {\tt firstCS}. Therefore the CP solutions are automatically converted to bit vectors $x$ and evaluated the product $x'Qx+constant$, where $constant$ is the bias that is not contained in the QUBO matrix, which only contains the coefficients of the quadratic and linear terms (given $b^2=b$ for any bit $b$). 
The tests on the problems $\mbox{Set}_i$ with $i\in\{1,2,3,6,7\}$ have shown that the modelling with QUBO is compatible with the solutions obtained classically and that those evaluate to the expected value (the makespan), which is thus proving that they fulfil all constraints modelled in the QUBO matrices and that bit vectors do exist that evaluate that low with our QUBO matrices.

In this work we were able to model an extented JSP using CP and as QUBO that is automatically converted to Ising Hamiltonian for adiabatic quantum computers or simulators. The model has been extended to incorporate the transport of workpieces through multiple AGVs.
The QUBO problems have been minimized on several solvers, including  Tabu Search, and adiabatic quantum computers. As a contrast, the original problem (not converted to QUBO) has been solved also with CP-based optimisation.

Quantum solvers are capable of finding similar results to those produced by the classical methods in small problem sizes. These solutions prove the QUBO models working but currently there is limited benefit for real life use cases. CP still performs better. As we increase the complexity of the JSP, i.e. increase the required number of qubits and qubit couplings, we observe a degrading of the quantum results versus the classical results. We believe the degrading occurs because of noisy qubit systems, limited resolution in implementing larger QUBO matrices and limited quantum coherence times in nowadays systems. 

\begin{figure}[ht]
    \centering
    \includegraphics[width=\linewidth]{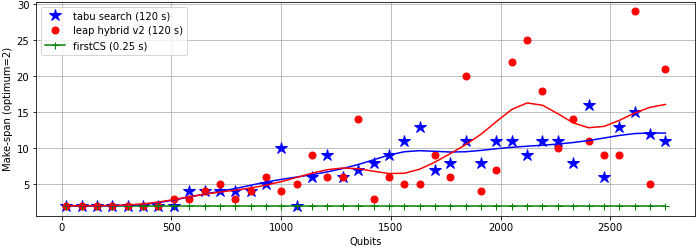}
    \vspace{-1.0em}
    \caption{Comparison of the solution quality for the same running time.}
    \label{fig:scalability}
\end{figure}

If the problem size grows into the order of several hundred qubits, the QUBO model needs to be decomposed and hybrid solver mechanisms are applied. We see hybrid approaches as best effort, since they rely strongly on classical optimization. The full details of hybrid classical solvers implemented in the cloud are known only to the manufacturers of the quantum computers so they cannot be discussed here. Our comparison showed a very similar performance of the local tabu search on 14 CPU cores and of the DWave hybrid in the cloud on a series of problems that differ only in the horizon and thus in the number of qubits (Figure \ref{fig:scalability}). The winner in this comparison is the CP-based approach, which delivers constantly within 250~ms the optimum makespan value.

The magnitude of compelling JSPs easily require the number of 5000, even 50.000 qubits. This requirement seems to be out of range of nowadays quantum (hybrid) approaches. We compared a JSP
of about 5000 qubits with various solvers and observe a clear advantage of classical optimization. However, we are curious to see progress of adiabatic quantum computing technology and we are ready to test future capabilities for real life JSPs in the 50.000 qubit regime.

The problem sizes that we are currently able to address are below business relevance and can be executed on classical computers with similar performance. 
It is uncertain right now, when the quantum computers will bring advantages for practical decision or optimization applications. Once that happens, it is likely that quantum computers will replace classical computers, on specialized problems like optimization problems, that have often exponential time complexity.


\printbibliography

%
%
%
%

\end{document}